\documentclass[runningheads]{llncs}
\usepackage{graphicx}
\usepackage{comment}
\usepackage{amsmath,amssymb} 
\usepackage{color}

\DeclareMathOperator{\E}{\mathbb{E}} 
\usepackage{algorithm, algorithmic}%
\usepackage{multirow}
\usepackage{hyperref}

\begin{document}
\pagestyle{headings}
\mainmatter
\def\ECCVSubNumber{2}  

\title{Next-Best View Policy for 3D Reconstruction} 

\titlerunning{Next-Best View Policy for 3D Reconstruction}
%
\author{Daryl Peralta\inst{1} \and
Joel Casimiro\inst{1} \and
Aldrin Michael Nilles\inst{1} \and
Justine Aletta Aguilar\inst{1} \and
Rowel Atienza\inst{1} \and
Rhandley Cajote\inst{1} }
\authorrunning{D. Peralta et al.}
%
\institute{University of the Philippines \\
Electrical and Electronics Engineering Institute \\
\email{\{daryl.peralta, joel.casimiro, aldrin.michael.nilles, justine.aguilar, rowel, rhandley.cajote\}@eee.upd.edu.ph}}

\maketitle

\begin{abstract}
    Manually selecting viewpoints or using commonly available flight planners like circular path for large-scale 3D reconstruction using drones often results in incomplete 3D models. Recent works have relied on hand-engineered heuristics such as information gain to select the Next-Best Views. In this work, we present a learning-based algorithm called \textbf{Scan-RL} to learn a Next-Best View (NBV) Policy. To train and evaluate the agent, we created \textbf{Houses3K}, a dataset of 3D house models. Our experiments show that using \textbf{Scan-RL}, the agent can scan houses with fewer number of steps and a shorter distance compared to our baseline circular path. Experimental results also demonstrate that a single NBV policy can be used to scan multiple houses including those that were not seen during training. The link to \textbf{Scan-RL}'s code is available at \url{https://github.com/darylperalta/ScanRL} and \textbf{Houses3K} dataset can be found at \url{https://github.com/darylperalta/Houses3K}.
\end{abstract}

\section{Introduction}

In recent years, there is an increased demand in 3D model applications including autonomous navigation, virtual and augmented reality, and 3D printing. Of particular interest in this work are 3D models of large infrastructure such as buildings and houses which can be used for construction monitoring, disaster risk management, and cultural heritage conservation.
	
Common methods in creating large 3D scenes include the use of color or depth images. Image-based methods use Structure from Motion (SfM) algorithms \cite{snavely2006photo,schonberger2016structure} that simultaneously estimate the camera poses and 3D structure from images. Depth-based methods such as \cite{izadi2011kinectfusion,dai2017bundlefusion} fuse depth images from different sensor positions to create the 3D model. These algorithms rely heavily on the quality of the viewpoints that were used. Lack of data in some parts of an object creates holes in the 3D reconstruction. A solution to this is to add more images from different viewpoints. However, using more images becomes computationally expensive and results to longer processing time \cite{heinly2015reconstructing}.

 In doing large-scale 3D reconstruction with drones, a pilot normally sets waypoints manually or uses commonly available planners like circular path. However, these methods often result in incomplete or low quality 3D models due to lack of data in occluded areas. Multiple flight missions are needed to complete a 3D model. This iterative process is time consuming and prohibitive due to the limited flight time of drones. 

The problem of minimizing the views required to cover an object for 3D reconstruction is known as the View Planning Problem \cite{kaba2017reinforcement}. This problem is also addressed by selecting the Next-Best View (NBV) which is the next sensor position that maximizes the information gain. These problems are widely studied because of their importance not only for 3D reconstruction but also for inspection tasks, surveillance, and mapping. 
    
    \begin{figure}[!t]
\begin{center}
   \includegraphics[width=0.75\textwidth]{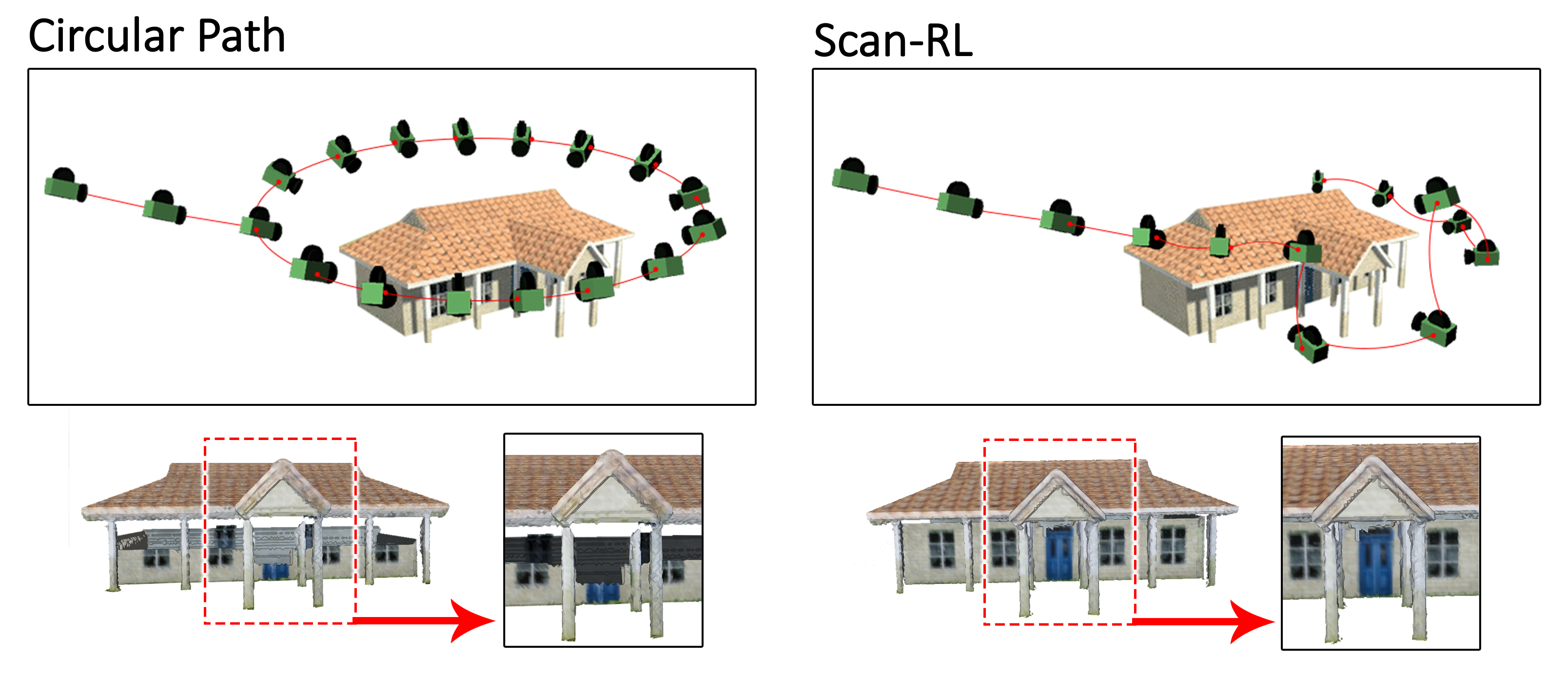}
\end{center}
	\caption{\textit{Upper left}: Viewpoints using a traditional circular path. \textit{Lower left}: Output 3D model using circular path with 87 $\%$ Surface Coverage. \textit{Upper right}: Viewpoints using \textit{Scan-RL} which was able to capture occluded regions under the roof. \textit{Lower right}: Output 3D model using \textit{Scan-RL} with 97 $\%$ Surface Coverage.}
    \label{fig:VP_overview}
\end{figure}

In this paper, an algorithm is proposed to answer the question: can an agent learn how to scan a house efficiently by determining the NBV from monocular images? Humans can do this task by looking at the house since we can identify occluded parts such as under the roof. We explore this idea and propose \textit{Scan-RL}, a learning-based algorithm to learn an NBV policy for 3D reconstruction. We cast the problem of NBV planning to a reinforcement learning setting. Unlike other methods, that rely on manually crafted criteria to select NBV, \textit{Scan-RL} trains a policy to choose NBV. Furthermore, \textit{Scan-RL} only needs images in making its decisions and does not need to track the current 3D model. Depth images and the reconstructed 3D model are only used during training.

Training \textit{Scan-RL} requires a dataset of textured 3D models of houses or buildings. To the best of our knowledge, there is no sufficiently large dataset of watertight 3D models of big structures suitable for our experiments. Most datasets focus on the interior like House3D \cite{wu2018building}. Thus, we created \textit{Houses3K}, a dataset made of 3,000 watertight and textured 3D house models. We present a modular approach to creating such dataset and a texture quality control process to ensure the quality of the models. While \textit{Scan-RL} was the motivation behind \textit{Houses3K}, it could also be used for other applications where 3D house models may be needed such as training geometric deep learning algorithms and creating realistic synthetic scenes for autonomous navigation of drones. 

Experiments were done in a synthetic environment to evaluate the algorithm. Results show that using \textit{Scan-RL}, an NBV policy can be trained to scan a house resulting in an optimized path. We show that the path using the NBV Policy is shorter than the commonly used circular path. A comparison of the circular path and the resulting path from \textit{Scan-RL} is shown in Figure \ref{fig:VP_overview}. Further experiments also show that a single policy can learn to scan multiple houses and be applied to houses not seen during training. 

	
	To summarize, our paper's main contributions are:
	
	\newcounter{qcounter}
\begin{list}{\arabic{qcounter})~}{\usecounter{qcounter}}
\item \textit{Scan-RL}, a learning-based algorithm to learn an NBV policy for 3D reconstruction for scanning in an optimal path. 
\item \textit{Houses3K}, a dataset of 3D house models that can be used for future works including view planning, geometric deep learning, and aerial robotics.

\end{list}

\section{Related Work}

    The challenge of view planning and active vision is widely studied in the field of robotics and computer vision. A survey of early approaches for sensor planning was done by Tarabanis et al. \cite{tarabanis1995survey}. Scott et al. \cite{scott2003view} presented a survey of more recent works in view planning. Our work is also related to active vision which deals with actively positioning the sensors or cameras to improve the quality of perception \cite{aloimonos1988active,chen2011active}.
    
    
    Approaches on planning views for data acquisition may be divided into two groups. The first group are those that tackle the view planning by reducing the views required to cover an object from a set of candidate views. These include \cite{smith2018aerial,kaba2017reinforcement,hepp2018plan3d}. The second group are those from the robotics community which aim to select the NBV in terms of information gain. These works include \cite{wenhardt2007active,potthast2014probabilistic,isler2016information,daudelin2017adaptable,song2018surface,kriegel2015efficient}. Our work is closer to NBV algorithms. \textit{Scan-RL} tries to train a policy that commands the drone where to position next given its current state to maximize improvement in the 3D reconstruction.
    
    For view planning, Smith et al. \cite{smith2018aerial} proposed heuristics to quantify the quality of candidate viewpoints. They also created a dataset and benchmark tool for path planning. Our method aims to learn a policy that will select the NBV for each step instead of simultaneous optimization of camera positions. 
    
    
       
    
    In NBV algorithms, a way to quantify information gain from each candidate viewpoint is needed. Delmerico et al. \cite{delmerico2018comparison} presented a comparative study of existing volumetric information gain metrics. Isler et al. \cite{isler2016information} proposed a way to quantify information gain from a candidate view using entropy contained in 3D voxels. Their algorithm chooses the candidate view with maximum information gain. Daudelin et al. \cite{daudelin2017adaptable} also quantifies entropy in voxels and proposed a new way of generating candidate poses based on the current object information. However, these works are limited to the resolution of voxels.

    
    
    

    Surface-based methods were also proposed \cite{kriegel2015efficient,song2018surface}. Kriegel et al. \cite{kriegel2015efficient} proposed to use the mesh representation of the 3D model together with a probabilistic voxelspace. Mesh holes and boundaries were detected for the NBV algorithm. Song et al. \cite{song2018surface} used truncated Signed Distance Field (TSDF), a surface representation. TSDF provides the information in improving the quality of the model based on confidence. Both algorithms track a volumetric representation for exploration and find boundaries between explored and unexplored voxels.
    
    
     Wu et al. \cite{wu2014quality} used points to estimate the NBV. The Poisson field from point cloud scans was used to identify the low quality areas and compute the next best view to improve the 3D reconstruction. This approach aims for completeness and quality of the 3D reconstruction. Huang et al. \cite{huang2018active} extended the work by introducing a fast MVS algorithm and applying the algorithm to drones. 
    
    All works presented previously relied on hand-engineered algorithms to compute the NBV. We aim to train a policy that will learn to select the NBV instead of manually proposing criteria for view quality. 
    
    Yang et al. \cite{yang2018active} used deep learning to predict the 3D model given the images coupled with reinforcement learning for a view planner. Their main contribution was the view planner which selects views to improve the 3D model predictions. However, the network only predicts 3D models of small objects with simple shapes and low resolution. Also, both the training of the 3D prediction network and the reward for the view planner rely on ground truth models. 
 
    Choudhury et al. \cite{choudhury2018data} proposed a learning-based approach to motion planning but not for 3D reconstruction. Kaba et al. \cite{kaba2017reinforcement} formulated a reinforcement learning approach to solve the view planning problem. However, their algorithm assumes a 3D model is available to minimize the number of views. Han et al. \cite{han2019deep} used a deep reinforcement learning algorithm for view planning as part of scene completion task. However, the policy was trained to maximize the accuracy of the depth inpainting task and was not trained to optimize the path. They also used the 3D point cloud as state. \textit{Scan-RL} aims to train a policy that optimizes the path using only monocular images as state. 
    

\section{Scan-RL: Learning a Next-Best View Policy}

\begin{figure*}[!h]
\begin{center}

   \includegraphics[width=0.75\textwidth]{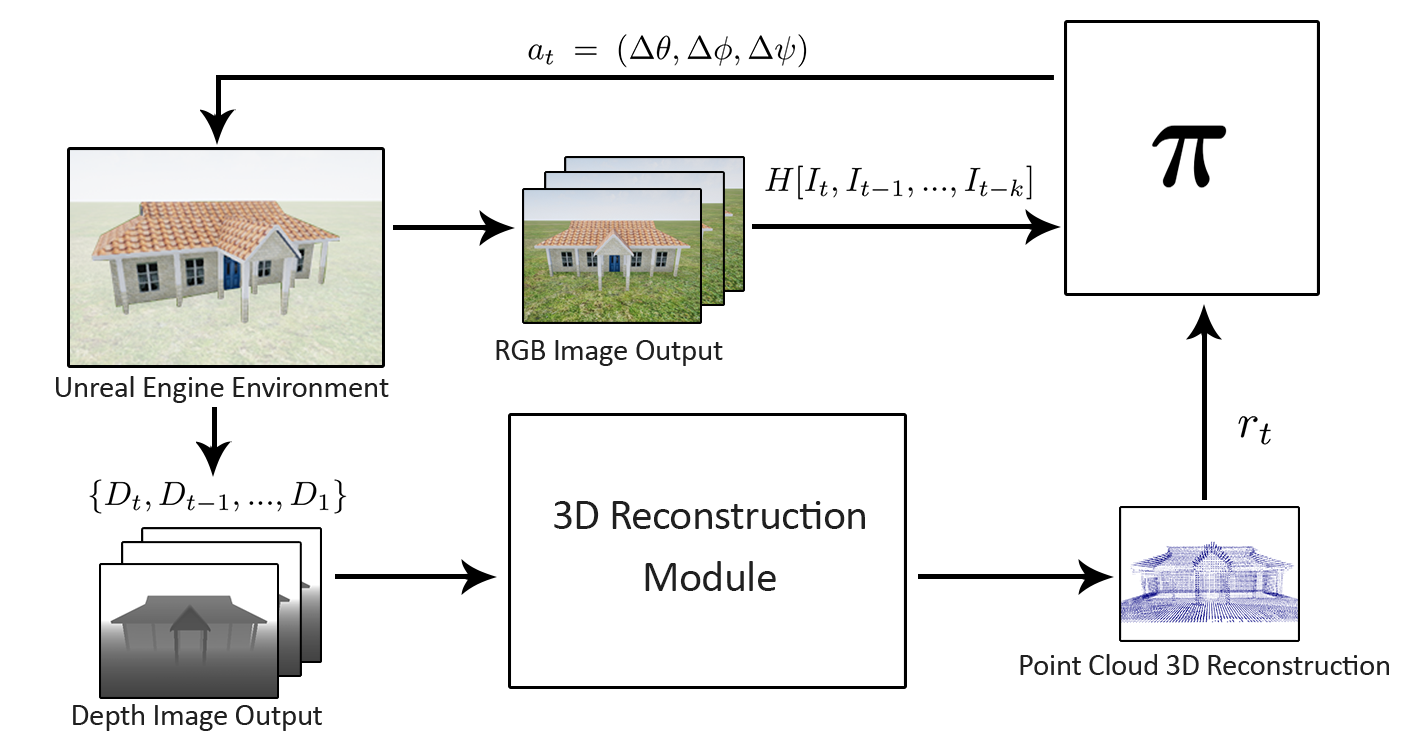}
\end{center}
	\caption{Block diagram of \textit{Scan-RL}. From the Unreal Engine environment, RGB images are rendered and used by the policy to generate the Next-Best View. Depth is rendered in the next view for the 3D Reconstruction module to create a point cloud 3D model. During training, rewards are extracted from the 3D model for the policy to learn scanning houses.}
    \label{fig:NBV_diagram}
\end{figure*}

A block diagram of \textit{Scan-RL} is shown in Figure \ref{fig:NBV_diagram}. The system is composed of two main components: the NBV policy $\pi$ and the 3D reconstruction module. The NBV policy $\pi$ selects the next pose to scan the target house based on images generated by the Unreal Engine\footnote{https://www.unrealengine.com/en-US/feed} environment. The 3D reconstruction module reconstructs the 3D point cloud model of the target structure using the depth images collected. Rewards are then extracted from the output point cloud.

\textit{Scan-RL} is modular. The NBV policy $\pi$ can be any policy trained using any reinforcement learning algorithm. The 3D reconstruction module can  be any 3D reconstruction algorithm that may utilize the monocular images, depth images, and camera poses from the synthetic scene. To make training feasible, a fast 3D reconstruction algorithm is needed. We utilized depth fusion algorithm based on Truncated Signed Distance Function (TSDF) representation introduced by Curless et al. \cite{curless1996volumetric}. TSDF is being used in recent depth-based reconstruction algorithms \cite{izadi2011kinectfusion,dai2017bundlefusion}. For the reinforcement learning algorithm, we used Deep Q-Network (DQN) \cite{mnih2015human} and Deep Deterministic Policy Gradient (DDPG) \cite{lillicrap2015continuous} which had success in high dimensional state space. 

Unreal Engine, a game engine, was used to create realistic synthetic scenes for our experiments. Unreal Engine allows controlling the camera positions and rendering color and depth images. We created our own synthetic game scenes where 3D house models from \textit{Houses3K} were used as shown in Figure \ref{fig:NBV_diagram}. To make the scene more realistic and feature-rich, we applied a grass texture for the ground. We built on top of Gym-unrealcv \cite{gymunrealcv2017} and UnrealCV \cite{qiu2016unrealcv} to implement the reinforcement learning algorithm and control the agent position in our synthetic environment.

We cast the NBV selection process as a Markov Decision Process (MDP). MDP is made of \((S,A,T,R,\gamma)\) where \(S\) is the state, \(A\) is the action, \(T\) is the transition probability, \(R\) is the reward and \(\gamma\) is the discount factor. We approximate an MDP process by making the state be the image captured by the camera at a viewpoint and action be the change in pose. We then applied reinforcement learning to find an optimal policy which we call NBV policy to select the next views.

\subsection{State Space}

In reinforcement learning, the state contains the necessary information for the agent to make its decision. Normally, robots and drones have multiple sensors like GPS, IMU, depth sensors, and cameras. To make our algorithm not dependent on the robot platform, we only used monocular images as state since cameras are commonly found in most drones. We refrained from using the 3D model as part of the state because it will require more computational resources. 

We define a preprocessing function $H$ shown in Eq. \ref{eq:statespace}. $H$ generates the state vector by concatenating the current image $I_t$ with previous $k$ frames, converting the color images to grayscale, and resizing them to a smaller resolution.

\begin{equation}\label{eq:statespace}
s_t = H[I_t, I_{t-1}, ...,  I_{t-k}]
\end{equation}


\subsection{Action Space}

Change in relative camera pose was used as action because we want our method to be independent of the dynamics of the robot platform. This also separates the low level control making it applicable to any robot which can measure its relative camera pose.



Camera pose was parameterized to \((\theta, \phi, \psi)\) \cite{su2015render}. \(\theta\) is the azimuth angle, \(\phi\) is the elevation angle and \(\psi\) is the distance from the object.  This assumes that the object is at the origin. Using this parameterization, the camera pose is reduced to three degrees of freedom.

We then define the discrete action space to be the increase or decrease in the camera poses resulting to six actions namely: increase \(\theta\), decrease \(\theta\), increase \(\phi\), decrease \(\phi\), increase \(\psi\) or decrease \(\psi\). The resolution for the change in camera pose is \((\pm 45 ^\circ, \pm 35 ^\circ, \pm 25\) units). For the continuous action space, we used the same relative camera pose \((\theta, \phi, \psi)\) but the change in angles is continuous resulting to \(a_t = (\Delta \theta, \Delta \phi, \Delta \psi)\).  Distance \(\psi\) is in Unreal Engine distance units.

\subsection{Reward Function}




The task to learn is selecting the NBV to completely scan an object. With this task, the reward should lead to a complete 3D model. To quantify completeness of the 3D model, we used surface coverage $C_s$ from \cite{isler2016information}.  Surface coverage \(C_s\) given by Eq. \ref{noeq:2} is the percentage of observed surface points \(N_{obs}\) over the total number of surface points \(N_{gt}\) in the ground truth model. \(N_{obs}\) are points in the ground truth model with a corresponding point in the output reconstruction whose distance is less than some threshold.



\begin{equation}\label{noeq:2}
C_s = \frac{{N_{obs}}}{{N_{gt}}}
\end{equation}

Reward \(r_t\) for each step is defined by Eq. \ref{noeq:3}. During steps where the terminal surface coverage is not achieved, reward is the change in surface coverage $k_c * \Delta C_s$ minus a negative penalty equal to $-2$ per step and a penalty proportional to distance $\Delta X$. The scaled change in surface coverage $k_c * \Delta C_s$ is introduced for the agent to maximize the surface coverage while the negative penalties are added to minimize the number of steps and distance. A reward of 100 is given when terminal surface coverage is achieved to emphasize the completion of the scanning task. The constants $k_c$ and $k_x$ were set to 1 and 0.02 respectively.

\begin{equation}\label{noeq:3}
 r_t =  
\begin{cases} 
      k_c * \Delta C_s - k_x * \Delta X - 2 & C_s\leq C_{s,terminal} \\
      100 & C_s > C_{s,terminal} \\
       
   \end{cases}
\end{equation}

To efficiently compute the reward $r_t$ and make training feasible, we needed a fast algorithm for the 3D reconstruction module. We implemented depth fusion algorithm \cite{curless1996volumetric} that generates a point cloud from all depth images $\{D_t,D_{t-1},...,D_1\}$ and camera poses. Depth-based algorithms are faster than image-based 3D reconstruction algorithms since depth information is already available. 



\subsection{Training}

Deep Q-Network (DQN) \cite{mnih2015human} was implemented for the deep reinforcement learning algorithm in the discrete action space. In DQN, a deep neural network is trained to maximize the optimal action-value function $Q^*$ expressed in Eq. \ref{noeq:4}. The optimal action-value function $Q^*$ is the maximum sum of rewards with discount factor $\gamma$ given state $s_t$ and action $a_t$ following a policy $\pi$.

\begin{equation}\label{noeq:4}
Q^* = \max\limits_{\pi}  \E(r_t + \gamma r_{t+1} + \gamma ^2 r_{t+2} + ...|s_{t}=s, a_t = a, \pi)
\end{equation}

For the continuous action space, we implemented an actor-critic algorithm Deep Deterministic Policy Gradient (DDPG) in \cite{lillicrap2015continuous}. DDPG is made of two neural networks namely actor and critic networks. Similar to DQN, the critic network in DDPG with parameters $\theta_{w}^Q$ is also trained to maximize Eq. \ref{noeq:4}. The actor network with parameters $\theta_{w}^\mu$ predicts the action that maximizes the expected reward. It is trained by applying the sampled policy gradient $\nabla_{\theta^\mu}J$ for a minibatch of $N$ transitions  $(s_j, a_j, r_j, s_{j+1})$ expressed in Eq. \ref{eq:actor-ddpg}.

\begin{equation}\label{eq:actor-ddpg}
\nabla_{\theta^\mu}J \approx  \frac{1}{N}\sum_{j}\nabla_aQ(s,a|\theta_{w}^Q)|_{s=s_j,a=\mu(s_j)}\nabla_{\theta_{w}^\mu}\mu(s|\theta_{w}^\mu)|s_j
\end{equation}


From the preprocessing function $H$, color images are converted to grayscale and are resized to $84\times84$. The preprocessing function $H$ then concatenates the previous $k$ frames which we set to 5 frames. This will result to a state vector of $84\times84\times6$. 



At every step, depth and pose are extracted from the synthetic scene. These are used in the 3D reconstruction module. Surface coverage is then computed from the output point and the reward $r_t$ is computed based on the reward function in Eq. \ref{noeq:3}. To compute the surface coverage, we generated the ground truth 3D point cloud by sampling 10,000 points from the ground truth mesh of \textit{Houses3K}. 

A house model is placed in the scene in each training episode. The episode terminates when the terminal surface coverage is achieved. When multiple models are being used during training, a model is drawn randomly from the dataset at the start of each episode. This can be done for the agent to learn features from different types of houses to perform the scanning task. Additional details about the training are presented in the supplementary material.


\begin{figure*}[!hbt]
\begin{center}
   \includegraphics[width=0.5\textwidth]{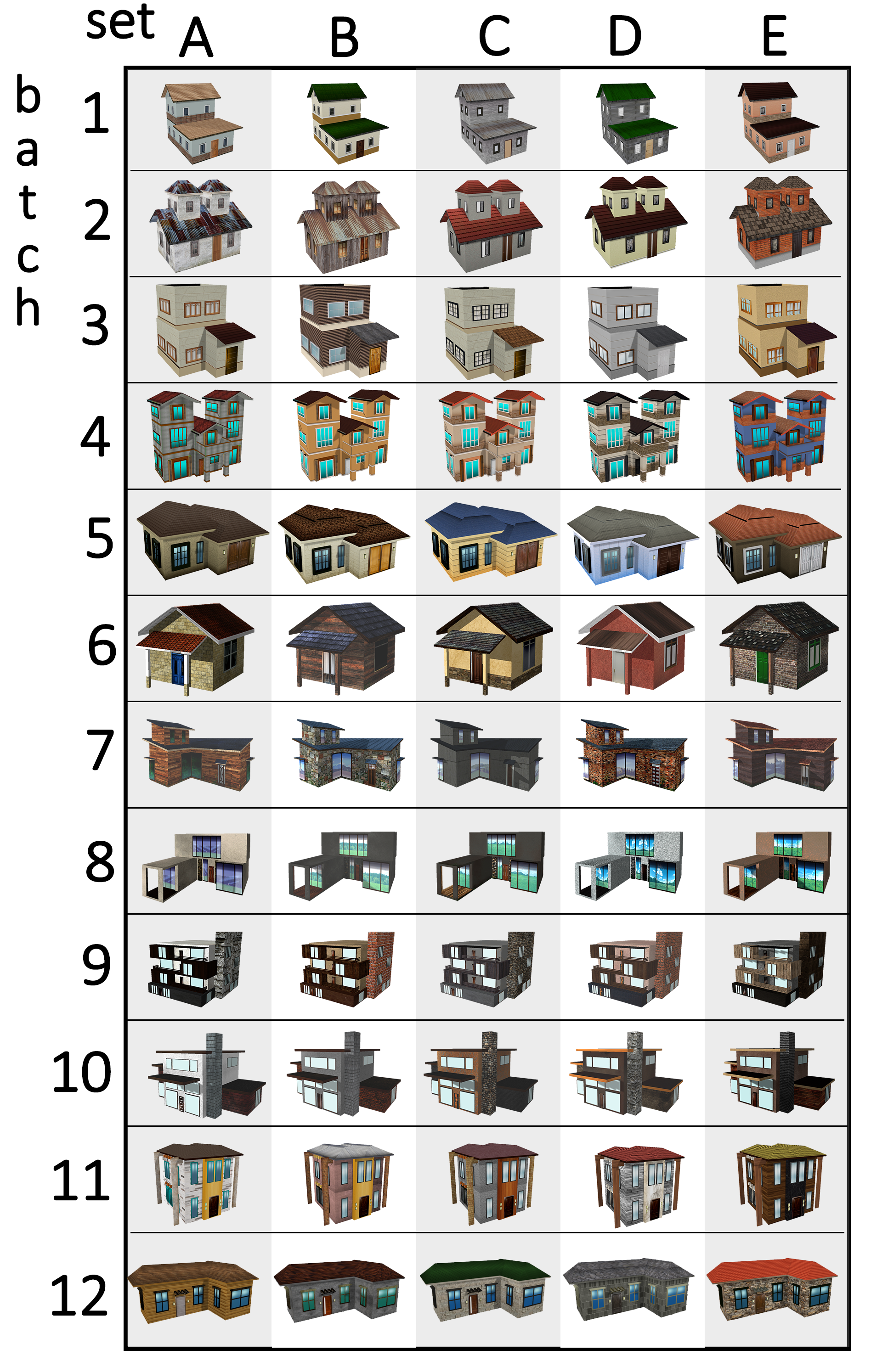}
\end{center}
	\caption{Sample houses for each set of \textit{Houses3K}. A dataset of 3000 3D house models grouped into 12 batches and 5 sets in each batch.}
    \label{fig:houses3k_table}
\end{figure*}

\subsection{Houses3K}

To train the policy, a dataset of textured 3D models of houses is needed. However, there is no suitable dataset that fits our need of 3D mesh focusing on the exterior of the house. Thus we created \textit{Houses3K}, a dataset consisting of 3,000 watertight and textured 3D house models.  

The dataset is divided into twelve batches, each containing 50 unique house geometries. For each batch, there are five different textures applied on the structures, multiplying the house count to 250 unique houses. Various architectural styles were adopted, from single-storey bungalow house type, up to a more contemporary and modern style. Figure \ref{fig:houses3k_table} shows sample houses from \textit{Houses3K}.

\begin{figure*}[!htb]
\begin{center}

   \includegraphics[width=0.9\textwidth]{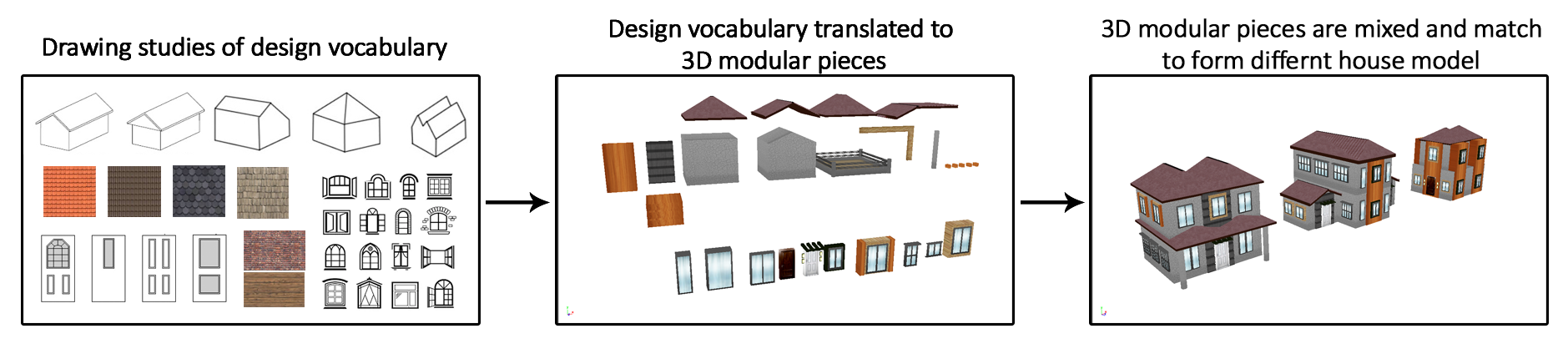}
\end{center}
	\caption{Modular approach to creating the \textit{Houses3K} dataset using 3D modular pieces.}
    \label{fig:modular-to-house}
\end{figure*}

We present a modular approach to creating the dataset illustrated in Figure \ref{fig:modular-to-house}. We created our own \textit{style vocabulary} which consisted of different house structure types, roof shapes, windows, and door styles, and even paints/textures for the roof and walls. Based on the \textit{style vocabulary}, modular 3D assets such as different styles of roofs, windows, doors, wall structures, and surface textures were created. These modular assets were then assembled in various combinations to make the different houses that comprise the dataset.


\begin{figure*}[!htb]
\begin{center}

   \includegraphics[width=0.95\textwidth]{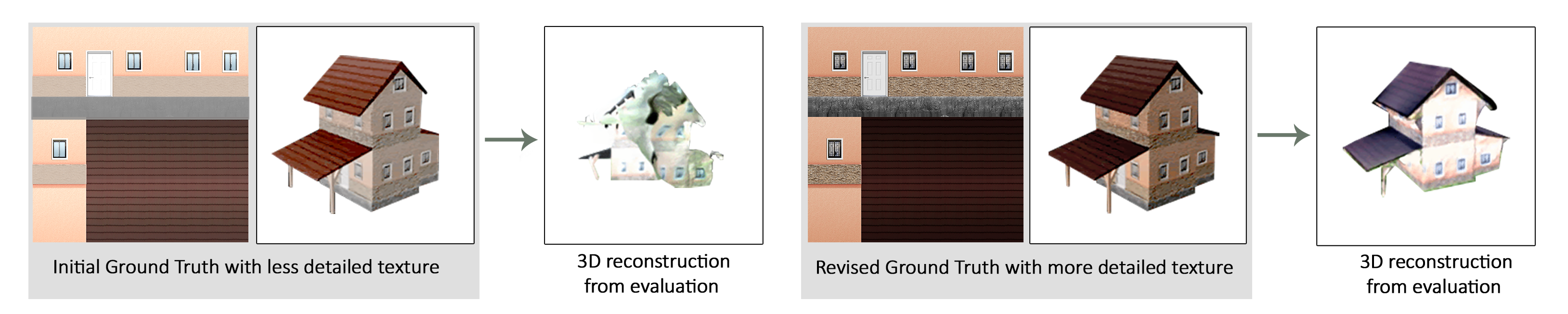}
\end{center}
	\caption{Texture quality control process. Rendered images from \textit{Houses3K} models were used as input to a multiple view image-based 3D reconstruction system to check if the models will be reconstructed.}
    \label{fig:texture-eval}
\end{figure*}

To ensure the quality of the textures, we implemented a texture quality control process illustrated in Figure \ref{fig:texture-eval}. Two representative models were sampled from each set and 108 images in a circular path with varying elevations were rendered at different viewpoints from these 3D models using Unreal Engine. These images were then used as input to an image-based 3D reconstruction system to generate a 3D model. If the model was not reconstructed properly as shown in the left side of Figure \ref{fig:texture-eval}, we redesigned the texture to make it feature rich. This process was done until the texture was detailed enough to reconstruct the model. 

\section{Experiments}

Three experiments were performed to evaluate \textit{Scan-RL}. The first experiment was the \textit{single house policy experiment} which aims to test if 
an agent can learn an NBV policy to scan a house efficiently using \textit{Scan-RL}.
The second experiment was the \textit{multiple houses single policy transfer experiment} which aims to test if a single policy can learn to scan multiple houses and be transferred to houses not seen during training. The last experiment was on Stanford Bunny\footnote{Available from Stanford University Computer Graphics Lab.}. This was conducted to test if \textit{Scan-RL} can be used for non-house objects and to compare our work with Isler et al. \cite{isler2016information}.






\subsection{Single House Policy Experiment}

We used the house model from batch 6 of \textit{Houses3K} shown in Figure \ref{fig:VP_overview} for this experiment. The house has self-occluded parts that will challenge the algorithm and will allow us to observe if the NBV policy can be used to efficiently scan houses. A terminal surface coverage of $96 \%$ was set during training. We implemented both discrete and continuous action space versions of \textit{Scan-RL}.

For the discrete action space, the number of distance $\psi$ levels were varied to two and three levels and azimuth $\theta$ was varied to $45.0 ^\circ$ and $22.5 ^\circ$ resolutions. Elevation $\phi$ was fixed to three levels. In the continuous action space, the maximum allowed distance from the origin was varied. The number of steps and the distance moved by the agent for the scanning task were measured and compared for all setups. 

\begin{figure}[!h]
  \centering
  \begin{minipage}[b]{0.3\textwidth}
    \includegraphics[width=\textwidth]{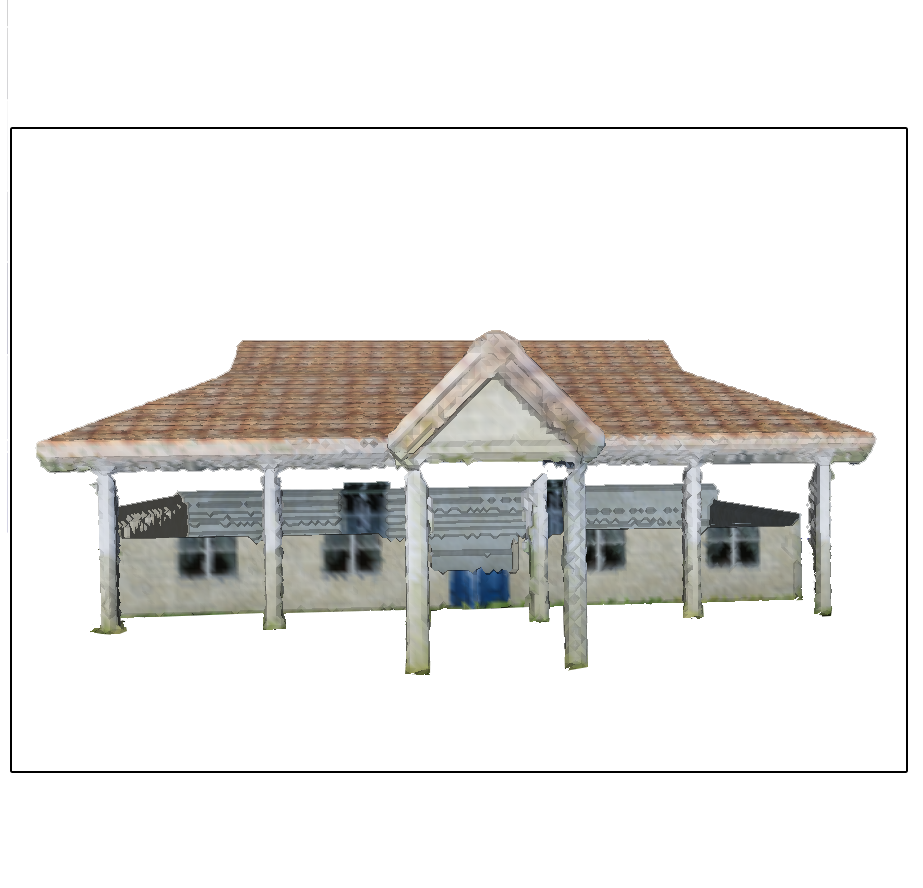}
    \caption{Circular path output 3D model. $87 \%$ surface coverage with 17 steps.}
    \label{fig:circular_out}
  \end{minipage}
  \hfill
  \begin{minipage}[b]{0.3\textwidth}
    \includegraphics[width=\textwidth]{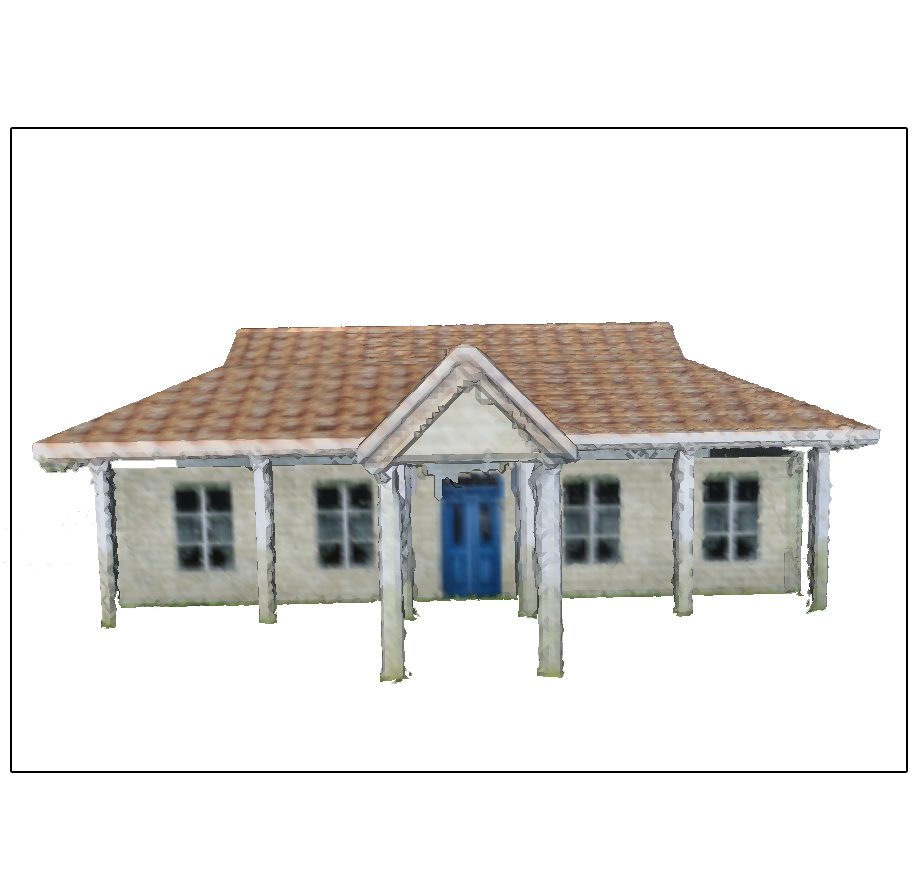}
    \caption{NBV policy output 3D model. $97 \%$ surface coverage with 13 steps.}
    \label{fig:vp_out}
  \end{minipage}
\end{figure}


We used the images and depth from the circular path and using the trained NBV policy from \textit{Scan-RL} to create the 3D models for each. The viewpoints selected for both methods are illustrated in Figure \ref{fig:VP_overview}. Output 3D models are shown in Figures \ref{fig:circular_out} and \ref{fig:vp_out}. Notice in Figure \ref{fig:circular_out} that the output 3D model has a large hole under the roof resulting to only $87 \%$ surface coverage with 17 steps. In Figure \ref{fig:vp_out}, the 3D model shows that parts under the roof were reconstructed resulting to $97 \%$ surface coverage using only 13 steps. This occurred since the policy was able to focus on the occluded regions under the roof. This show that the NBV policy learned to maximize the surface coverage to create a more complete 3D reconstruction while minimizing the number of steps. 

To further evaluate \textit{Scan-RL}, three types of circular paths with varying elevations were implemented for the discrete action space. The first type named \textit{Circ 1} path starts with the agent at the farthest distance level and at $45.0 ^\circ$ elevation. The agent moves up to the highest elevation level, moves closer to the object which is $100$ units from the origin, circulates around the object, and then moves to the middle and bottom elevation levels. \textit{Circ 2} path starts with the same position, the agent then moves straight to the nearest allowed distance which is $100$ units from the origin and circulates around the object as shown in Figure \ref{fig:VP_overview}. It then moves to the bottom then to the top elevation. \textit{Circ 3} path is similar to \textit{Circ 1} but starts with the bottom elevation. For the continuous action space setup, the range of action $a_{t}$ is  ($\pm45.0 ^\circ,\pm35.0 ^\circ, \pm25$ units).  A similar circular path baseline that chooses the maximum value in the action range was implemented for the continuous action space. 

\begin{figure}[!htb]
	\centering
    \includegraphics[width=0.6\textwidth]{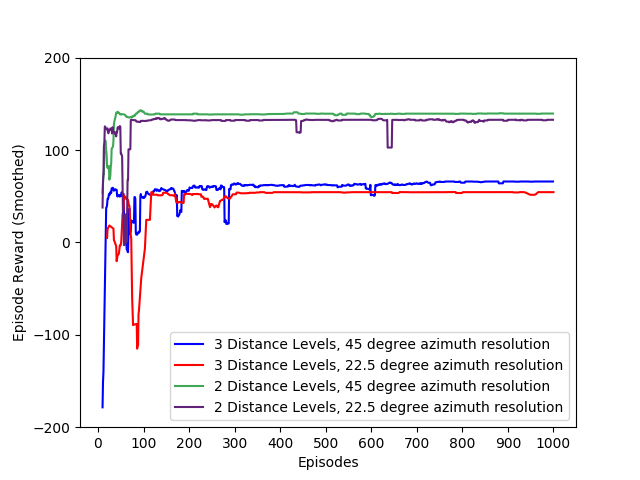}
    \caption{\textit{Scan-RL} performance during training for the different action space setups.}
    \label{single_house_train}
\end{figure}

Table \ref{table:single} shows results for the discrete action space with varying azimuth resolution and distance levels. In all setups, using \textit{Scan-RL}'s NBV policy resulted to fewer steps and shorter distance compared to the baseline circular path. Table \ref{table:cont} shows results for the continuous action space for two different maximum distance allowed from the origin. In the continuous setups, \textit{Scan-RL}'s NBV policy was able to optimize the steps but not the distance. This occurred because both are being optimized and it is possible to have fewer steps but longer distance. In Figure \ref{single_house_train}, the cumulative reward of \textit{Scan-RL} during training is shown. It can be observed that for all setups, the policy learned to maximize the reward throughout the training.

\begin{table*}
\small
\caption{Single House Policy Experiment in discrete action space. Number of steps and distance covered for $96 \%$ terminal coverage were compared with different baselines.}
\label{table:single}
\vskip 0.15in
\begin{center}
\begin{scriptsize}
\begin{sc}\begin{tabular}{|c|c| c| c| c| c | c| c| c| c|}
\hline
Distance & Azimuth & \multicolumn{4}{c |}{Steps} & \multicolumn{4}{c |}{Distance covered (units)} \\ 
\cline{3-10} Levels & Resolution & Circ 1 & Circ 2 & Circ 3 & Scan-RL  & Circ 1 & Circ 2 & Circ 3 & Scan-RL \\

    \hline
      2 & 45.0$^\circ$ & 23 & 12 & 8 & \textbf{7}   & 1087.30 & 662.87 & 510.62 & \textbf{410.13}  \\
    \hline
      2 & 22.5$^\circ$ & 44 & 23 & 13 & \textbf{12} & 1162.93 & 702.12 & 492.60  & \textbf{416.81} \\
    \hline
      3 & 45.0$^\circ$ & 24 & 13 & 9 & \textbf{8}  & 1097.54 & 687.87 & 520.85 & \textbf{435.13} \\
    \hline
      3 & 22.5$^\circ$ &  43 & 24 & 18 & \textbf{15} &1082.13 & 727.12 & 641.38 & \textbf{533.34} \\
    \hline
\end{tabular}
\end{sc}
\end{scriptsize}
\end{center}
\vskip -0.1in
\end{table*}

\begin{table*}
\small
\caption{Single House Policy Experiment in continuous action space. Number of steps and distance covered for $96 \%$ terminal coverage were compared with different baselines.}
\label{table:cont}
\vskip 0.15in
\begin{center}
\begin{scriptsize}
\begin{sc}\begin{tabular}{|c |c |c |c |c |c |c |c |c |}
\hline
Max Allowed & \multicolumn{4}{|c |}{Steps} & \multicolumn{4}{c|}{Distance covered (units)} \\
\cline{2-9} Distance (units) &  Circ 1 & Circ 2 & Circ 3 & Scan-RL & Circ 1 & Circ 2 & Circ 3 & Scan-RL \\ \hline
 {125}  & 23 & 12 & 8 & \textbf{7} & 1087.30 & 662.87 & \textbf{510.62} & 520.08 \\ \hline
 {150}  & 24 & 13 & 9 & \textbf{7} & 1097.54 & 687.87 & \textbf{520.85} & 550.80 \\ \hline
\end{tabular}
\end{sc}
\end{scriptsize}
\end{center}
\vskip -0.1in
\end{table*}


These results show that the agent was able to learn an NBV policy to scan the house and achieve the terminal surface coverage while minimizing steps and distance travelled by the agent.  The agent learned to move forward, move around the house, and go down for the occluded parts as shown in Figure \ref{fig:VP_overview} resulting to an optimized path. 


These results also show that it is indeed possible to scan a house efficiently using \textit{Scan-RL} without manually crafting criteria for NBV. The trained NBV policy uses only images as state vector. It means that the policy learned all the necessary information including the concept of relative position from the images. 

Being able to train an NBV policy that optimizes the path is important for large-scale 3D reconstruction using drones with limited flight. It also has potential applications to other tasks like inspection and exploration. 

\subsection{Multiple Houses Single Policy Experiment}

In this experiment, we implemented the discrete action space version of \textit{Scan-RL}. \textit{Houses3K} models were split into train and test sets per batch. House models in the train set were used for training a single policy to scan multiple houses. Models in the test set were used to test if the trained NBV policy can scan the unseen test models. We implemented two kinds of split namely \textit{random} and \textit{geometry} split. Random split randomly splits the houses to $90 \%$ for training and $10 \%$ per batch. \textit{Geometry} split also splits with the same ratio but makes sure that the geometry in the train set is not in the test set.

For each batch, a single policy was trained using \textit{Scan-RL} discrete space.  Hyperparameters were held constant for all batches. Solved house models in \textit{Houses3K} are defined to be models where the terminal coverage was achieved within 50 steps. A solved model means that the agent was not stuck during the process. For each batch, we also ran the circular path baseline which we ran for 27 steps since this is the number of steps required to cover all the possible viewpoints at the closest distance level.

For Random Split, the terminal surface coverage for all batches were set to $96 \%$ except for batches 8 and 10 which were set to $87 \%$ due to the models' complex geometry. For Geometry Split, the terminal surface coverages for all batches were also $96 \%$ except for batches 7, 8 and 10 which were set to $87 \%$. Table \ref{table:multiple_houses} presents the ratio of the number of solved houses using \textit{Scan-RL} over the total number of houses used for each batch. In this experiment, we are interested in the number of houses that can be solved so the number of steps were not included. It is also expected that the circular path will have a higher score since it will cover all the views in the closest distance level.

Results using the train set show that \textit{Scan-RL} can train a single NBV policy to scan multiple houses in \textit{Houses3K} as seen in the high number of houses solved except for the more complex batches 7 and 9. Results from the test set show that for most batches, the NBV policy was able to scan the unseen houses. Failure cases were mainly caused by the complex geometries which may require more degrees of freedom in the camera pose. 




\begin{table}
\small
\caption{Ratio of solved houses using \textit{Scan-RL} over the total number of houses used for each \textit{Houses3K} batch.}
\label{table:multiple_houses}
\vskip 0.15in
\begin{center}
\begin{scriptsize}
\begin{sc}\begin{tabular}{|l | c | c | c | c | c | c | c | c | c|}
\hline

        &  \multicolumn{4}{c |}{Random Split} &  \multicolumn{4}{c |}{Geometry Split} \\
      \cline{2-9}  &  \multicolumn{2}{c |}{Train Set} & \multicolumn{2}{c |}{Test Set} & \multicolumn{2}{c |}{Train Set} &
      \multicolumn{2}{c |}{Test Set} \\
    \cline{2-9} Batch  & Scan-RL &Circ & Scan-RL &Circ & Scan-RL & Circ & Scan-RL & Circ \\
     \hline 1 & 217/225 & 225/225 & 23/25 & 24/25 & 218/225 & 224/225 & 25/25 & 25/25  \\
     \hline 2 & 224/225 & 225/225 & 25/25 & 25/25 & 224/225 & 225/225 & 24/25 & 25/25  \\
     \hline 3 & 182/225 & 186/225 & 18/25 & 22/25 & 203/225 & 190/225 & 16/25 & 20/25\\
     \hline 4 & 204/225 & 206/225 & 24/25 & 24/25 & 193/225 & 210/225 & 17/25 & 20/25 \\
     \hline 5 & 222/225 & 225/225 & 25/25 & 25/25 & 214/225 & 225/225 & 25/25 & 25/25 \\
     \hline 6 & 213/225 & 215/225 & 25/25 &25/25 & 204/225 & 220/225 & 20/25 & 20/25 \\
     \hline 7 & 76/210 & 137/210 & 10/20 & 13/20 & 197/210 & 190/210 & 17/20 & 20/20 \\
     \hline 8 & 183/225 & 189/225 & 18/25 & 21/25 & 131/225 & 185/225 & 16/25 & 25/25 \\
     \hline 9 & 155/225 & 191/225 & 13/25 &19/25 & 145/225 & 190/225 & 11/25 & 20/25 \\
     \hline 10 & 194/225 & 214/225  & 14/25 & 20/25 & 214/225 & 211/225 & 20/25 & 25/25 \\
     \hline 11 & 191/225 & 225/225 & 18/25 & 25/25 & 220/225 & 225/225 & 25/25 & 25/25 \\
     \hline 12 & 195/225 & 189/225 & 18/25 & 18/25 & 184/225 & 190/225 & 19/25 & 25/25 \\
     \hline      
\end{tabular}
\end{sc}
\end{scriptsize}
\end{center}
\vskip -0.1in
\end{table}

\subsection{Non-House Target Model Experiment}

We replicated the setup of Isler et al. \cite{isler2016information} in our synthetic environment. The main differences in the setups are that we used the depth rendered from Unreal Engine for the 3D reconstruction module and that we generated our own ground truth point cloud with 10,000 points from the ground truth mesh. To have the same set of viewpoints to the one used by \cite{isler2016information}, we used the $22.5 ^\circ$ azimuth resolution of our discrete action space setup. We also set the depth sensor range such that we get the maximum surface coverage of 90 \%. Training was the same with the single house policy experiment including the hyperparameters.

\begin{figure}[!htb]
	\centering
    \includegraphics[width=0.55\textwidth]{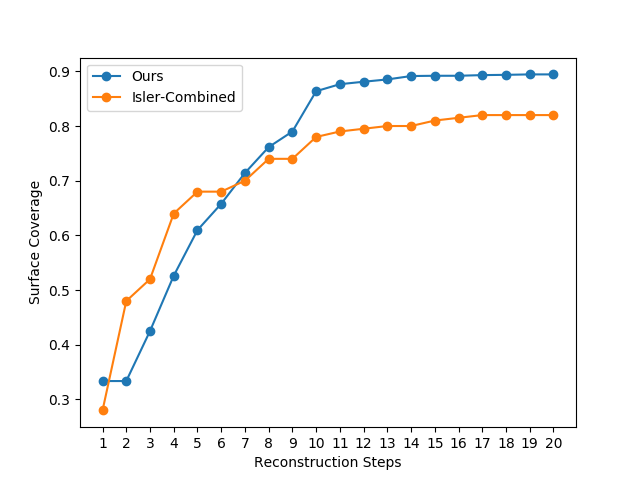}
    \caption{Reconstruction of Stanford Bunny. Surface Coverage per step of our method is compared with Isler \cite{isler2016information}.}
    \label{bunny_sc_plot}
\end{figure}

The plot of surface coverage per reconstruction step on Stanford Bunny is shown in Figure \ref{bunny_sc_plot}. Data from Isler et al. \cite{isler2016information} were compared with the performance of \textit{Scan-RL}'s trained NBV policy. Both methods were able to increase the surface coverage per step and converge to a high surface coverage. This experiment shows that \textit{Scan-RL} is not only limited to houses. 


\section{Conclusion}
We presented \textit{Scan-RL}, a learning-based algorithm to train an NBV policy for 3D reconstruction inspired by how humans scan an object. To train and evaluate \textit{Scan-RL}, we created \textit{Houses3K}, a dataset of 3,000 watertight and textured 3D house models which was built modularly and with a texture control quality process. Results in the single house policy experiment show that \textit{Scan-RL} was able to achieve $96 \%$ terminal surface coverage in fewer steps and shorter distance than the baseline circular path. Results in the multiple houses single policy evaluation show that the trained NBV policy for each batch in \textit{Houses3K}, can learn to scan multiple houses one at a time and can also be transferred to scan houses not seen during training. Future works can look into deploying the NBV policy trained in the synthetic environment to a drone for real world 3D reconstruction.  

\section{Acknowldegements}

This work was funded by CHED-PCARI Project IIID-2016-005 (AIRSCAN Project).

\clearpage
%
%
\bibliographystyle{splncs04}
\bibliography{egbib}
\end{document}